# Artificial Eye for the Blind

Abhinav Benagi , Dhanyatha Narayan , Charith Rage , A Sushmitha
Computer Science Enginnering , M S Ramaiah Institute Of Technology, India

## Abstract

Visual Impaired humans cannot perceive their environment and navigate like normal Humans do , which results in reduced mobility.They also fear the society and are often treated as incable submissive section of society. In the light of making a blind person overcome his fear and step out to the  society independently and confidently, we have added a few features to the blind stick enabling the blind to be able to detect gender,age and basic actions done by a person through which he can react to his opposite party in an appropriate manner.

This project is useful when it comes to blind people being able to detect facial expressions in order to respond to environmental situations precisely.It also helps them detect the age and gender of the person to respond accordingly. It also helps them interpret text from images without relying on the third party for help.For security purposes, it helps in detecting the actions carried out by the person in the surroundings and taking appropriate actions against the situation. If this project is implemented appropriately, it can help blind people in carrying out a wide range of everyday activities effortlessly which adds on a lot of scope for the juvenile generation.

For facial expression detection, AI uses a lot of non-verbal functionalities such as facial expressions, body language and gestures. Whereas, to determine the age and gender it uses a broad dataset which includes a significant amount of pictures of men and women contributing to the dataset which in turn helps in accurately determining the age and gender of the person.Text to speech conversion is done by storing many different font and text images in patterns. It uses matching algorithms to distinguish between text images, character by character.In conclusion, the blind stick accommodates multiple features each of which can be accessed with separate functionalities such as, one functionality for action and object detection , one for image to speech conversion , one for facial recognition.

# Table of content

# Chapter 1 - Introduction

Globally, Around 2.2 Billion people don't have the capability to see and 90% of them coming from low-income countries tell us that, it is a need of the hour for an easily accessible, economically viable and ethically appropriate equipment for the specially abled. Unknown environments are especially challenging as it is difficult for a person to detect the objects in the surroundings or read any sign boards, understand the behavior of the human or navigate themselves to the destination.

Blindness is one of the most misunderstood types of disability ,as many people believe that a blind person cannot do their work or live normally. Millions of them are in India and are facing troubles in their daily lives as we don't have the proper equipment for them. Adding on, common able-people are often judgemental about the troubles faced by the visually impaired which makes them hesitant to face the dominant world.The objective of this project is to ensure that the visually impaired will now emerge to be more confident and feel more empowered as compared to  other sections of the population. With the implementation of this project, the blind can now be less dependent on their current environment and people.They can now get first hand access to technology and make best use of it.

There are several devices and aids for the blind that are already available But, all of them focus on a specific functionality. Our project comes with a solution that can accommodate provisions for multiple functionalities during the same computational period making it stand out amongst the existing devices. Braille is one of the places the specially abled can comfortably read in different languages too. But, it too is not available everywhere across the globe thus pushing them towards a means for reading and recognizing text. This section of the society, being more prone to fatalities on the road accidents need a feasible device through which they can stay afar from dangers.

Hence, in this project we have included features like object detection,image to text conversion and then to speech,gender and age detection and ultimately action detection. Furthermore it can also be converted to speech so that the blind person can hear using his earphone. When a blind person needs to know directions by reading the sign boards or when he has to read a non-braille book he can just click the picture of it using the feature of image to text conversion where, the captured image can be analyzed and the English text which is present can be recognised and by using text to speech conversion feature the text is read out and can be  understood by the blind. A sensor connected to the blind stick beeps when the object is in the range of specified distance.

Action detection is the task of identifying when a person in an image or video performs atomic actions such as walking ,bending,clapping,falling etc. This helps a blind person to evaluate the situation and react accordingly.It is vital for any species to know what is going on and how the world revolves around them. When you are visually impaired, chances of your other senses becoming extremely sharp are really high. Most blind people would automatically develop sharp hearing senses which is a good sign of empowerment but it's not necessary that only improved senses can help a blind get through any difficult situation. Although it's not impossible to deal with situations without action detection it is sometimes really helpful if the blind can detect what actions are being done by what people around him. For instance let us consider there's a man putting up an act in an open space environment, it would be helpful for the blind to know what's happening around to be aware of any dangerous acts occurring around him.

The methodology used in object and action detection is, capturing a real time snapshot of the present situation and using tensor flow modules and libraries to detect what activity the respective person is doing and what object he/she is holding.This can be done using video detection also instead of capturing snapshots by feeding the program with a pre recorded video. Alternatively, real time video detection was not chosen since the microcomputer cannot handle high GPU power. The mentioned features are implemented via raspberry pi microcontroller.

Object recognition is a classical problem in computer vision, determining if the image contains a specific object.so we chose this problem upon color recognition, shape recognition etc.This feature identifies objects present in real time more accurately as the dataset contains images in many different contexts .It can perform better at identifying objects in various environment.Object detection bounds boxes around these detected objects which allows us to identify the coordinates of each object exactly .

Optical character recognition(OCR) is an optical scanner, also a specialized software ,used to extract text from images . Google Text to Speech(gTTS) library in python is used,which converts text into audio and allows us to save this audio as a mp3 file. The speech can be delivered at various speeds and can be customizable based on the user's grasping level .

Age and gender detection is implemented using deep learning concept to accurately identify age and gender of a person from a single image of a face.It is very difficult to accurately guess an exact age from a single image because of factors like makeup,lightning,obstruction an facial expressions so we make this a classification problem instead of regression. For all of us it is important that we are addressed appropriately, with respect to age and gender. A lady wouldn't want to be addressed as male and the older section of the society would not like to be disrespected. Therefore it is important for the visually impaired to be able to identify the age and gender of an individual so that he/she can fit into the society without much trouble.

Our Blind stick project implements multiple functionalities different from a normal blind stick which only beeps on the detection of an object in front of it, which is exactly the cause for why we think this project has a wide scope in future. Only when we walk in their shoes will we know the casualties and difficulties faced by them. In this project we have tried our level best in understanding the problems faced by the blind and tried implementing a solution for it.

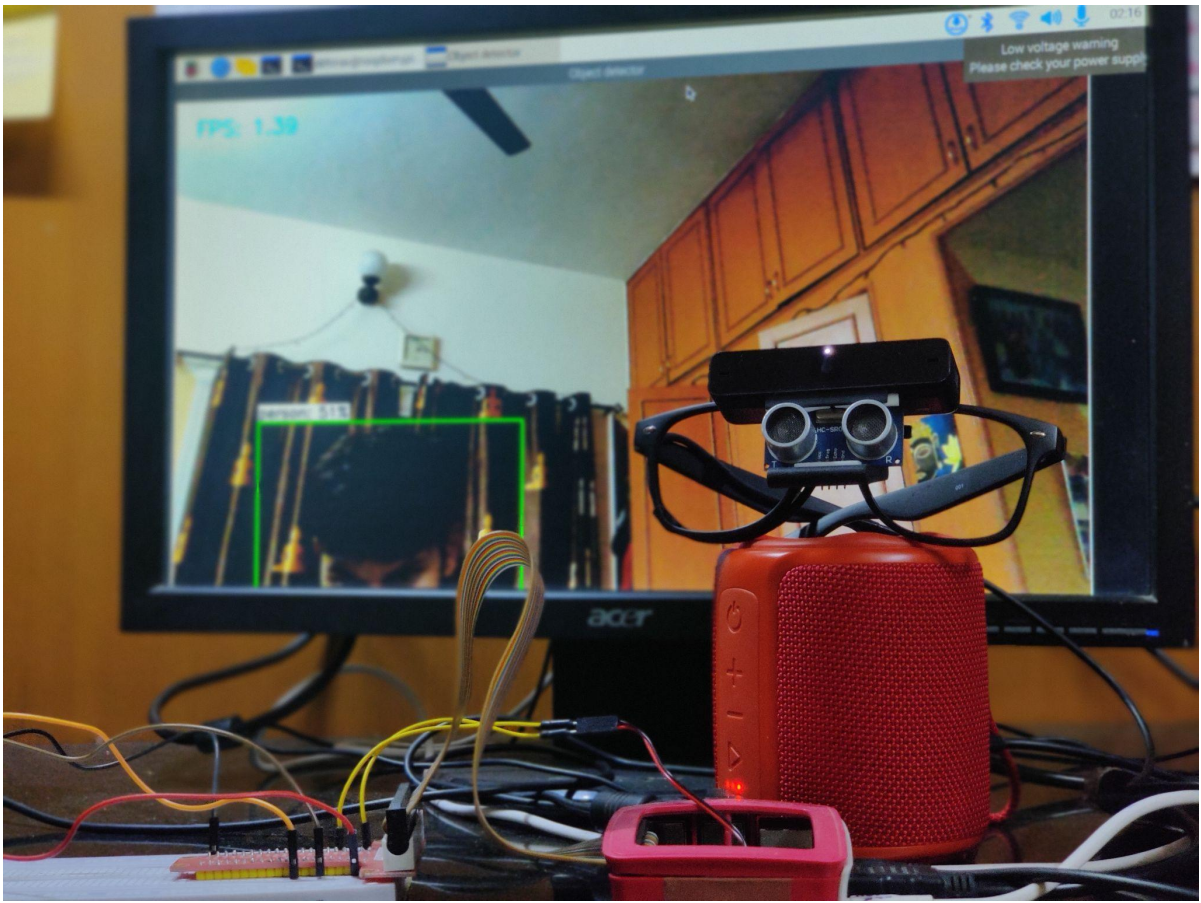

Fig 1: Artificial Eye

# Chapter 2 - Literature Review

There are many advantages of the advanced driving assistance system(ADAS) for a better driving experience such as this system operates on radars, LiDARs for object detection and therefore providing our model high speed, low cost and lower power consumption. The proposed system used in this paper for object detection using YOLOv5 which helps us reduce the occlusion issue. We also study the services and benefits of darknet CNN and its 53 convolutional layers which when stacked onto the original architecture summing up to 106 layers. Darknet also acts as a backbone to the YOLOv5 algorithm as proposed in [1].

The recent developments of object detection using deep learning by building convolutional neural networks(CNN) was observed. We also learn the evaluation of tensor flow object detection framework for robust detection of traffic light [2]. It also tells us that faster RCNN delivers 97.015%, which outperformed a single shot multibox detector(SSD) by 38.806% for a model trained with a set of images.A case study about traffic lights in Malaysia and the images that were collected and stored as a dataset was also demonstrated. Detection of eyes and the sobriety of the person is done to determine whether the person is under the influence of alcohol. Safety in self-driving cars is improved using the model [2].

We understand that CNNs are one of the best algorithms to process the image content and have shown immense improvements in the image content segmentation, classification, detection and retrieval related tasks from [3]. CNN also has a feature to capitalize on spatial or temporal correlation in data. The nonlinearity generated in the different patterns of activation of different responses therefore gives a different semantic understanding for images. The ability to learn good representation from raw pixels without exhaustive processing, hierarchical learning, automatic feature extraction, multi-tasking and weight sharing makes it stand out amongst other algorithms. Backpropagation algorithm helps in the learning of the model by manipulating the change in the weights according to its respective target. Thus, we understand that deep architectures have an edge over shallow architectures due to its human level performance.

YOLO (you only look once) is a Powerful algorithm used for image processing through which objects can be detected with great precision and accuracy. A single convolutional network parallelly predicts bounding boxes and class probabilities for those boxes. The algorithm trains full images and optimizes our overall performance. This  model has vast benefits over traditional methods of object detection. It could be compared with other powerful algorithms such as tensor flow with respect to speed and accuracy. Its speed is extremely fast since detection is framed as linear regression and a simple pipeline is laid out [4]. Base network runs at 45 frames per second with no batch processing on a Titan X GPU and the fast version runs at more than 150 fps. This tells us that we can process streaming video in real-time with less than 25 milliseconds of latency[4]. We increase the precision by two times.

There are multiple sensing modalities like RGB cameras, depth cameras, inertial sensors etc through which actions of humans are determined. Fusion of the decisions obtained from two different modalities make the model more robust. Actions sometimes can be categorized into interest and non-interest actions from a very vast action stream is a primary necessity that has to be satisfied in an action detection model. There are several practical examples in the modern day world where actions dictate many tasks in our life. Gesture detection features are being incorporated into mobiles, TV's and any other automation where there is a need for human intervention, automations like these play a critical role in [5]. CNN(convolutional neural network) and Long short term networks(LSTM) based fusion systems are used to detect actions of interest from the stream.

Tesseract OCR Engine makes use of Long Short Term Memory (LSTM), a part of RCNN's[6]. It is suitable in recognizing larger portions of text data rather than single characters. Errors occurred are reduced during character recognition. It is difficult for visually impaired people to read textual information. The Blind have to make use of Braille to read. Instead it would be an easier task if they could simply listen to the audio version of the text. This application is a viable alternative to convert textual data to audio format. Google Text-To-Speech API facilitates this need.[6]

It is vital for the visually impaired to identify the activities performed by the people around them. This can be done through various convolutional models and tensorflow is one among them.The research [7] builds a human action recognition system based on a single image or a video captured. The TensorFlow Deep Learning models are developed using human keypoints generated by OpenPose. Four classifiers are considered: Neural Network, Random Forest, K-Nearest Neighbor (KNN), and Support Vector Machine (SVM) Classifiers.. The models' input layer is 50 points from x and y coordinates of 25 keypoints from OpenPose, and the output layer is the numerical representation of various human action labels which like hand waves, planks, running, sitting, hiding etc [7].

It is necessary for a person with visual aid to identify the gender and age in order to address the opposite party respectfully. Gender and age play a significant role in interpersonal interactions among people who live in communities. Image enhancement used in [8], is the process of improving an image so that the resultant image is of higher quality and can be used by other applications. The image is divided into a limited number of objects in order to solve the problem, this is called Segmentation. Due to the accuracy of its classification technique, deep learning techniques are a variety of tasks such as classification, feature extraction, object recognition, and so on. It helps in gender and age prediction.In the CNN model used in [8], first the face is extracted from a webcam image before proceeding with the implementation. The OpenCV library in Python is used to accomplish this. Haar feature-based cascade classifiers canbe used for detecting objects

Description, Feature extraction and gender classification is done sequentially in [9]. The input image is taken, then the central line location and various organs like eyes, noses, mouth are found. Feature extraction is done after locating all the components mentioned above with the help of PCA(principal component analysis). PCA helps us to predict, remove redundancies and compress the compiled data. It also tends to find a M dimensional subspace from a face image that is represented as a two dimensional N by N array of intensity values.The age of the person is determined with the help of a face space which is found out by computing the euclidean distance of feature points in two faces. Representation of data in columns of matrices and computing the covariance matrix help us to find the eigen vector and therefore allows us to determine the gender with accuracy.NNC(nearest neighbor classification) and KNN(kth nearest neighbor) classifiers are used to correctly classify the faces and determine the gender of the person.

Optical Character Recognition (OCR) is a branch of AI which is used to detect and extract characters from scanned documents or images and convert them to editable form. We then convert this extracted text to speech making it understandable to the visually impaired. Traditional methods of OCR used CNN's but they are complex and usually suitable for single characters. These methods also had a higher number of errors and less precision.[10]

# Chapter 3 – Implementation

The main backbone of our Artificial Eye model is the Raspberry pi3 which is connected to the webcam ,ultrasonic proximity sensor, speaker and we also run all our software models i.e object detection, Optical Character recognition, google text to speech conversion and the Mycroft voice assistance model.

At first the ultrasonic proximity sensor will be measuring the distance between itself and any obstacle in front of it .When the Proximity sensor detects any obstacle in front within its specified range, the blind person will hear an audio prompt about an obstacle in his way at a certain distance. At this time the Webcam will capture an image in front of it and the Object detection model and the Optical Character Recognition model will begin to run on the Raspberry pi. The image captured is first sent through the Tesseract OCR module to detect any texts in the image and then through the Object detection model to detect the objects in front of the blind person. The text and the object detected are conveyed to the blind person by converting the texts to speech by using the gTTS module.

Along with the above mentioned process going on there will be an active MYCROFT voice assistant model which can be used to interact with the blind person. The blind person can ask about the weather , daily news , any information on the internet ,etc.

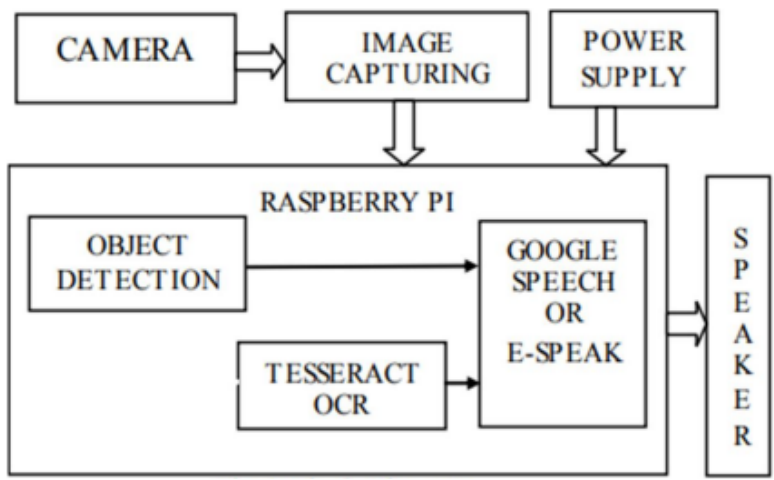

Fig 2: general working diagram

**Proximity Sensor**

We are using the Ultrasonic SR04 sensor as our proximity sensor.Ultrasonic sensors calculate's the obstales distance by emitting ultrasonic sound waves and converting those waves into electrical signals. The ultrasonic sensors have a range of up to 40-300cm with a great response time of up to 50 milliseconds to 200 milliseconds. This Proximity sensor's Vcc,gnd,Trig,Echo pins are connected to the Raspberry pi 3 b+ GPIO pins . We have used a python file in the raspberry pi for the working of this ultrasonic sensor.

The ultrasonic sensor keeps transmitting signals continuously , when an obstacle comes in front of it the receiver receives the signals and the obstacle is detected. Once the obstacle is detected the python script in the raspberry pi converts the distance into an audio format telling the blind person the distance between himself and the obstacle. We have used the Google Text to Speech api to convert the string to an audio output.

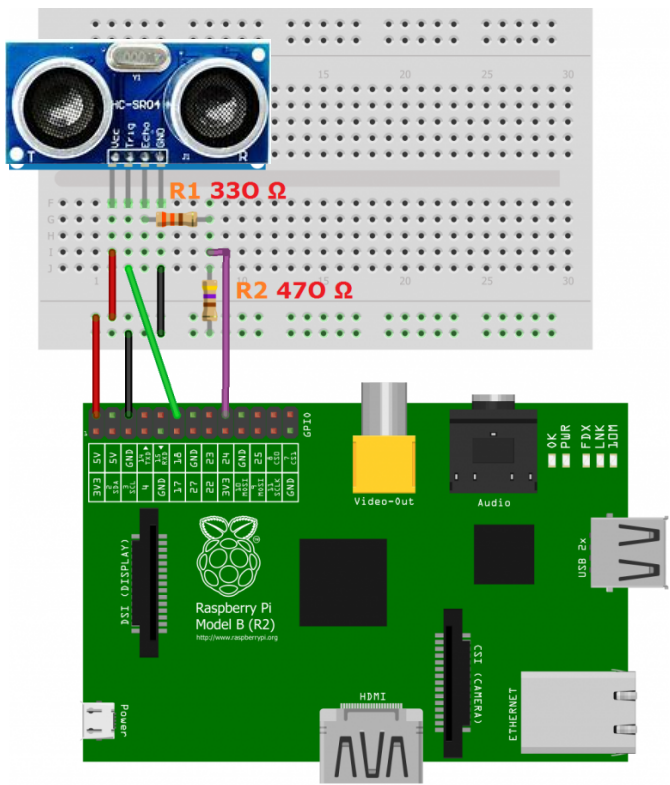

Fig 3. SR04 ultrasonic sensor connected to the raspberry pi

## Object detection and Optical character recognition

TensorFlow is a free and open-source software library for machine learning and artificial intelligence.We are using the tensor flow libraries for obstacle detection and optical character recognition.As we are using the Raspberry pi3 to process these deep neural network model results have shown that tensfor flow lite object detection model works much more efficiently and faster than the yolo algorithm.

In our project we have done the real time object detection using mobileNET_SSD algorithm from the TensorFlow lite framework. The TensorFlow lite is pre-trained on the COCO dataset.When the camera detects the object the class outputs which are detected are successfully converted into real time speech signal using the gTTS package of google in python library. So at this point our Artificial Eye model is able to detect any object with in the COCO dataset and convert it into an Audio format for the blind person to hear. As of now The coco dataset are trained only on 80 classes, for our future works we are trying to expand the dataset and train our own model to make it more accurate and effecient .

We will expand our model by customizing our own data for facial recognition of known people and gender recognition. We have also planned to implement the action detection model to convey the blind person about the action being performed around his environment.

## MYCROFT

Mycroft is a additional feature which we are implementing to make our project a business model. The blind people can't see television ,ther only source of entertainment in audio. Mycroft is a open source,Artificial Intelligent voice assistant which can be easily integrated with a Raspberry pi. The blind person can easily interact with this model through his day asking about the weather, date ,time , hear some local news or listen to a music of his choice

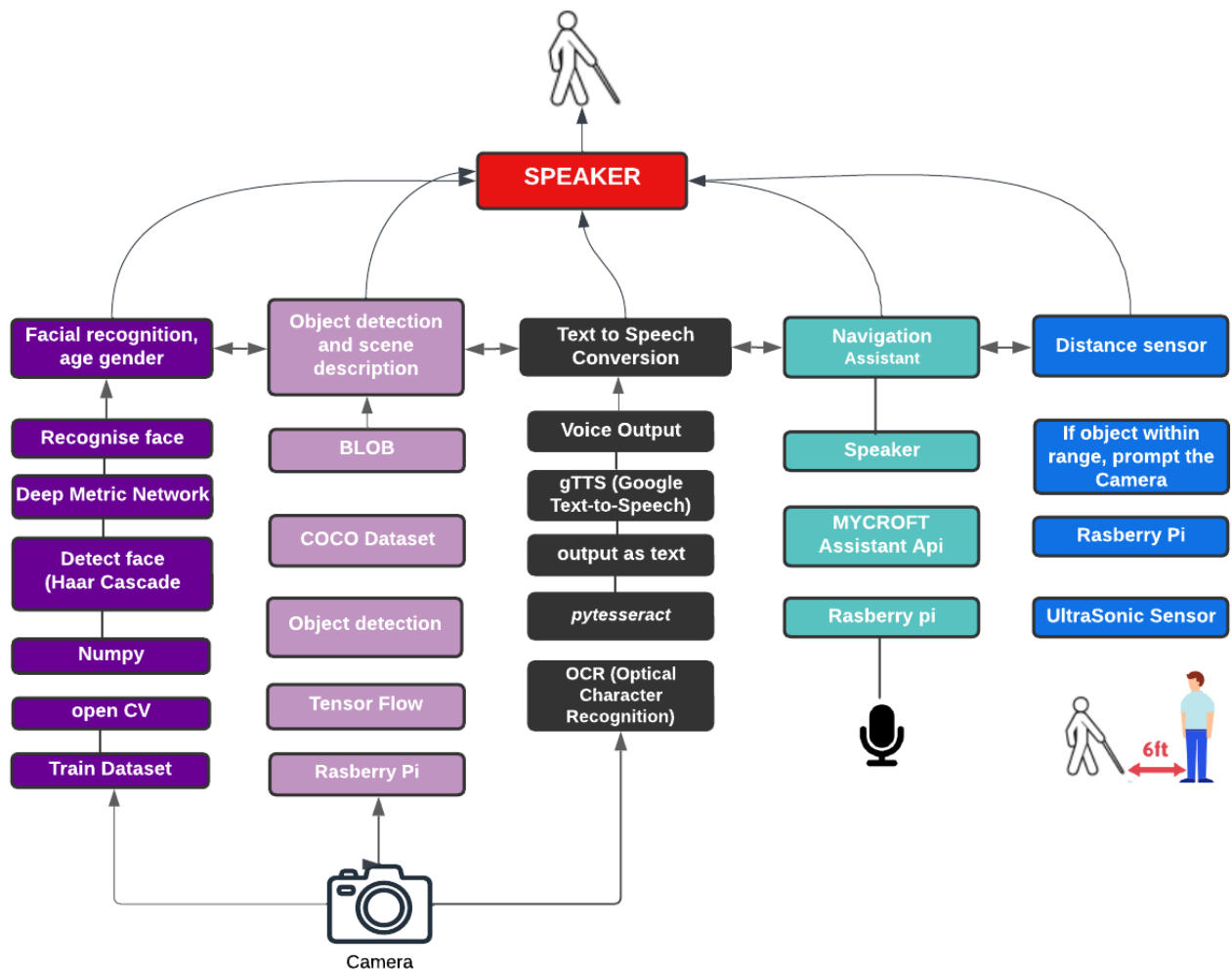

Fig 4: Tech Stack of the Artificial eye model

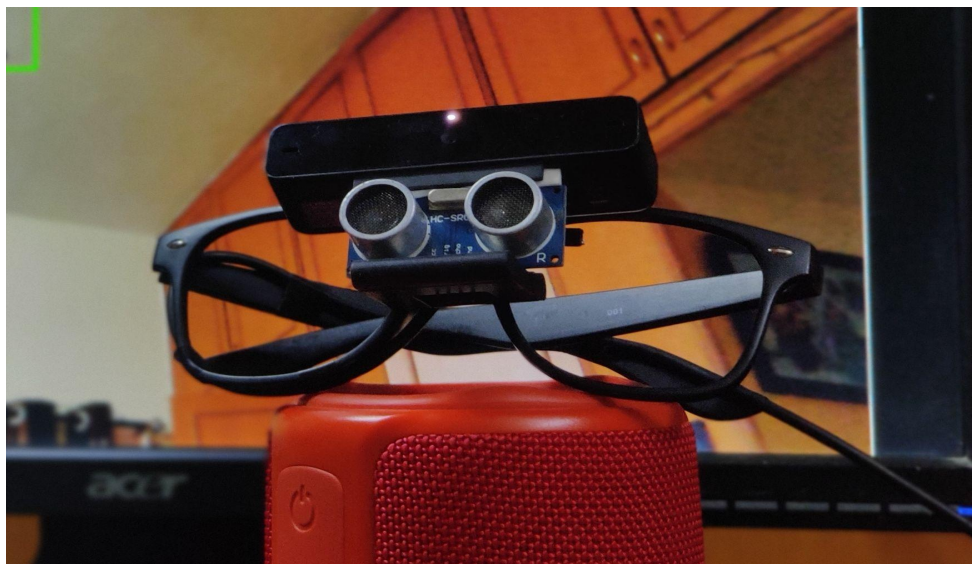

Fig 5 : The image of the Smart eye

# Chapter 4 – Evaluation Results

## Results drawn from the Proximity Sensor

The time taken by the raspberry pi to execute the Ultra sonic sensor are as follows

| Distance(cm) | Time to execute |
|---|---|
| 8.7 | 0.0037789 |
| 32 | 0.003605 |
| 61 | 0.00524 |
| 62 | 0.00587 |
| 65 | 0.00560188 |
| 77 | 0.00597 |
| 97 | 0.007149 |
| 118 | 0.00804 |
| 142 | 0.01002 |
| 161 | 0.0161 |

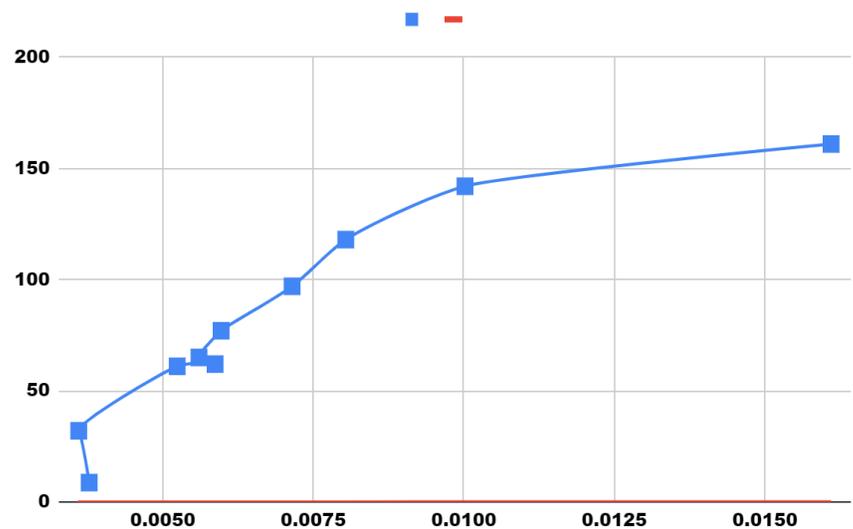

Fig 6: Graph plotting distance vs time take to execute by ultrasonic sensor

The Average response time of a human being is about 200milli seconds i.e upto 0.2 seconds but our ultrasonic sensor works much faster than the human eye with an average response time of 0.007137478 seconds or 0.7 milliseconds.
We can clearly say that our obstacle detection model is precise and accurate.

```
High Performance MPEG 1.0/2.0/2.5 Audio Player for Layer 1, 2, and 3.
Version 0.3.2-1 (2012/03/25). Written and copyrights by Joe Drew,
now maintained by Nanakos Chrysostomos and others.
Uses code from various people. See 'README' for more!
THIS SOFTWARE COMES WITH ABSOLUTELY NO WARRANTY! USE AT YOUR OWN RISK!

Playing MPEG stream from audio1.mp3 ...
MPEG 2.0 layer III, 32 kbit/s, 24000 Hz mono

[0:01] Decoding of audio1.mp3 finished.
distance measurement in progress
waiting....
Measure  Distance = 53.4 cm
time taken to execute 0.004654884338378906
High Performance MPEG 1.0/2.0/2.5 Audio Player for Layer 1, 2, and 3.
Version 0.3.2-1 (2012/03/25). Written and copyrights by Joe Drew,
now maintained by Nanakos Chrysostomos and others.
Uses code from various people. See 'README' for more!
THIS SOFTWARE COMES WITH ABSOLUTELY NO WARRANTY! USE AT YOUR OWN RISK!

Playing MPEG stream from audio1.mp3 ...
MPEG 2.0 layer III, 32 kbit/s, 24000 Hz mono

[0:05] Decoding of audio1.mp3 finished.
```

Fig7 :  output of Ultrasonic   sensor

# Comparative analysis of the Object Detection model

Case 1: Implementing  object detection on Rasberry Pi

Here we have done a complete performance analysis among most of the deep neutral network models for object detection. Deep neural network models like faster_RCNN , mobilenet-ssd,yolo v2, yolo v3,yolo v4 ,yolo v5 have been compared. We have compared these algorithms not only based on the mAP( mean Average Precision) but also GFLOPS (Giga Floating Point Operations Per Second )and MPARAMS (model parameters).

Most of the them only look for the results of only  mAP but in this case as we are implementing this on raspberry pi we also have to consider to the computing speed of the it  for which it is necessary to see the GFLOPS parameter.

What is GFLOPS?
GFLOPS stands for Giga_Floating_Point_Operations_Per_Second . It is a unit of measurementthat measures the performance of a floating point unit of a computer .Gigaflops measure the number of billions of floating-point calculations a processor can perform per second, and it directly tells us the  computing power of a processor.

As we are using a raspberry pi which does not have a powerful cpu to handle the processing of the top notch CNN models like YOLO v4 , we have to consider a balance between with mAP and GFLOPS in such a way that we neither compromise too much on the accuracy by using lesser GFLOPS nor select an Algorithm with high accuracy with higher GFLOPS .

Hence we have made a comparative analysis of the following models and come to a conclusion that MobileNet_SSD works the best for us.

| | GFlops | MParams | mAP | Source framework |
|---|---|---|---|---|
| faster_rcnn_inception_resnet_v2_ | 30.687 | 13.307 | 52.4 | TensorFlow* |
| faster_rcnn_inception_v2_coco | 30.687 | 13.307 | 40 | TensorFlow* |
| Faster R-CNN Resnet-50 | 57.203 | 29.162 | 42.87 | TensorFlow* |
| mobilenet-ssd | 2.316 | 5.783 | 79.8377 | Tensorflow* |
| ssd_mobilenet_v1_coco | 2.494 | 6.807 | 23.3212 | TensorFlow* |
| YOLO v2 Tiny | 5.424 | 11.229 | 27.34 | Keras* |
| YOLO v2 | 63.03 | 50.95 | 27.34 | Keras* |

| YOLO v3 Tiny | 5.582 | 8.848 | 35.9 | Keras* |
|---|---|---|---|---|
| YOLO v4 | 128.608 | 64.33 | 71.17 | Keras* |
| YOLO v3 | 65.984 | 61.922 | 62.27% | Keras* |
| YOLO V5s | 17 | 7.3 | 55.4 | Keras* |
| YOLO V5lite | 2.42 | 1.62 | 41.3 | Keras* |

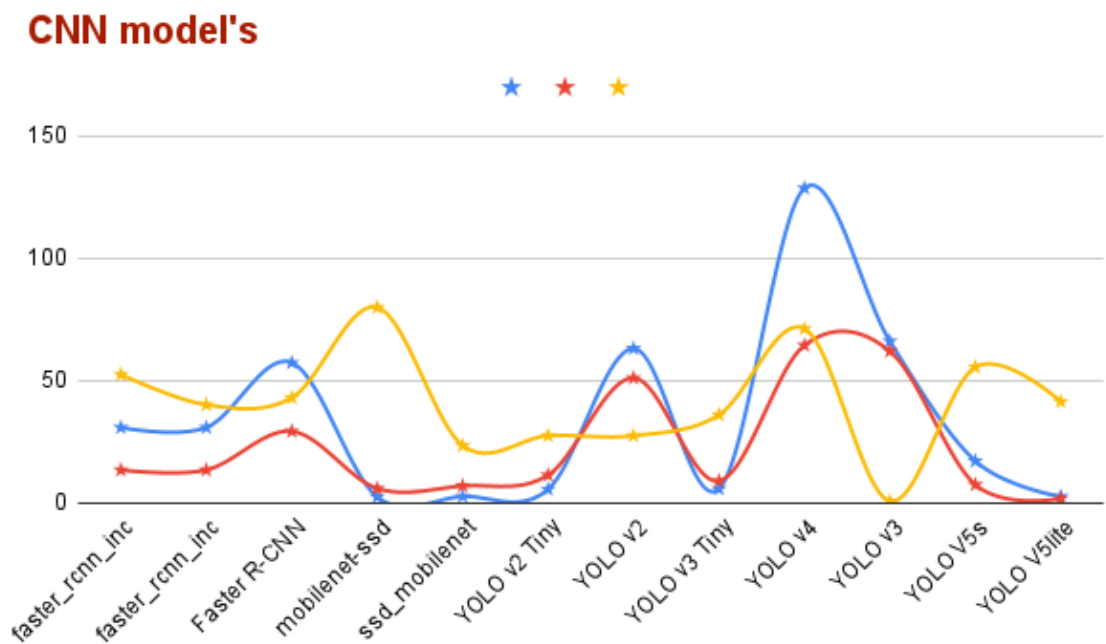

Fig 8 : Graph Ploting the Gflops,Mparams,mAP of different CNN models

Hence From this plotted graph it is very clear that the mobilenet_SSD uses the least GFLOPS with maximum mAP .Therefore we have used the TensorFlow Lite Framework which uses the mobileNet_SSD  to achieve our object detection on the raspberry Pi

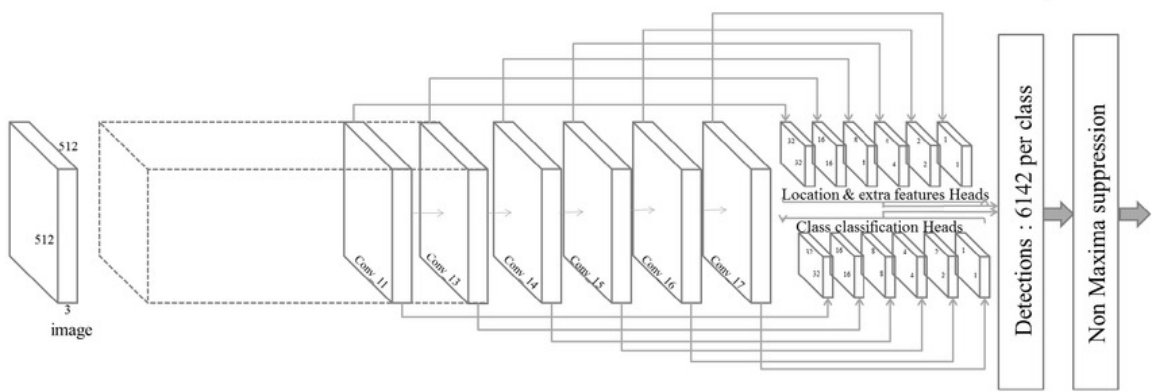

Fig 9:  pictorial representation of MobileNet V1 based SSD architecture patter

CASE 2 : Implementing Object detection on powerful CPU's

Here, in this case we compare YOLO v5, YOLO v4 and YOLO v3's performance with respect to accuracy, precision, error, speed and many other aspects.With regard to this project RASPBERRY PI has been used and accordingly YOLO v5 was an efficient algorithm which was light weight with size only 1.7M (int8) and 3.3M (fp16). It can reach 10+ FPS on the Raspberry Pi 3B when the input size is 320×320 Bytes.

COMPARISON OF RESULTS:

| ID | Model | Input_size | Flops | Params | Size (M) | Map@0.5 | Map@.5:0.95 |
|----|-------|-----------|-------|--------|----------|---------|-------------|
| 001 | yolo-fastest | 320×320 | 0.25G | 0.35M | 1.4 | 24.4 | - |
| 002 | nanodet-m | 320×320 | 0.72G | 0.95M | 1.8 | - | 20.6 |
| 003 | yolo-fastest-xl | 320×320 | 0.72G | 0.92M | 3.5 | 34.3 | - |
| 004 | yolov5-lite | 320×320 | 1.43G | 1.62M | 3.3 | 36.2 | 20.8 |
| 005 | yolov3-tiny | 416×416 | 6.96G | 6.06M | 23.0 | 33.1 | 16.6 |
| 006 | yolov4-tiny | 416×416 | 5.62G | 8.86M | 33.7 | 40.2 | 21.7 |
| 007 | nanodet-m | 416×416 | 1.2G | 0.95M | 1.8 | - | 23.5 |
| 008 | yolov5-lite | 416×416 | 2.42G | 1.62M | 3.3 | 41.3 | 24.4 |
| 009 | yolov5-lite | 640×640 | 2.42G | 1.62M | 3.3 | 45.7 | 27.1 |
| 010 | yolov5s | 640×640 | 17.0G | 7.3M | 14.2 | 55.4 | 36.7 |

Fig 10

mAP stands for mean average precision - mAP@0.5 means to say that  the mAP calculated at IOU threshold 0.5.
Map@.5:0.95 means average mAP over different IoU thresholds, from 0.5 to 0.95, step of 0.05

What is the IOU threshold?

Intersection over Union, a value used in object detection to measure the overlap of a predicted versus actual bounding box for an object. The closer the predicted bounding box values are to the actual bounding box values the greater the intersection, and the greater the IoU value, on the other hand, If the difference between the predicted and actual bounding box is less, then lesser is IOU.

COMPARISON ON DIFFERENT PLATFORMS:

| Equipment | Computing backend | System | Framework | Input | Speed(our) | Speed(yolov5s) |
|---|---|---|---|---|---|---|
| Inter | @i5-10210U | window(x86) | 640×640 | torch-cpu | 112ms | 179ms |
| Nvidia | @RTX 2080Ti | Linux(x86) | 640×640 | torch-gpu | 11ms | 13ms |
| Raspberrypi 4B | @ARM Cortex-A72 | Linux(arm64) | 320×320 | ncnn | 97ms | 371ms |

Fig 11

WHY IS YOLOv5 RECOGNISED AS THE BEST ALGORITHM?

*Inference speed:*The inference speed is measured with frames per second (FPS), namely the average iterations per second, which can show how fast the model can handle an input. The higher the inference speed, the faster, the better is the performance.YOLOv5 has higher inference speed which is why it is better than other YOLO models.

*Detection of small or far away objects:* Other models such as YOLO v3,v4 are not good when it comes to the accuracy of objects that are far away and of the objects that are small in size. YOLO v5 has a comparatively good accuracy when it comes to detecting objects in such a category.

*Little to no overlapping boxes*: YOLOv5 has great performance when it comes to separating bounding boxes for various objects at different distances apart from each other. Various other algorithms have a lot of overlapping issues with regard to bounding boxes which makes it difficult for detection of objects that are very near to each other.

*Speed*: This is a major advantage with the YOLOv5 model since it's an engine they developed to run sparse models on CPU. This is faster when compared to other GPU's and respective YOLO models.

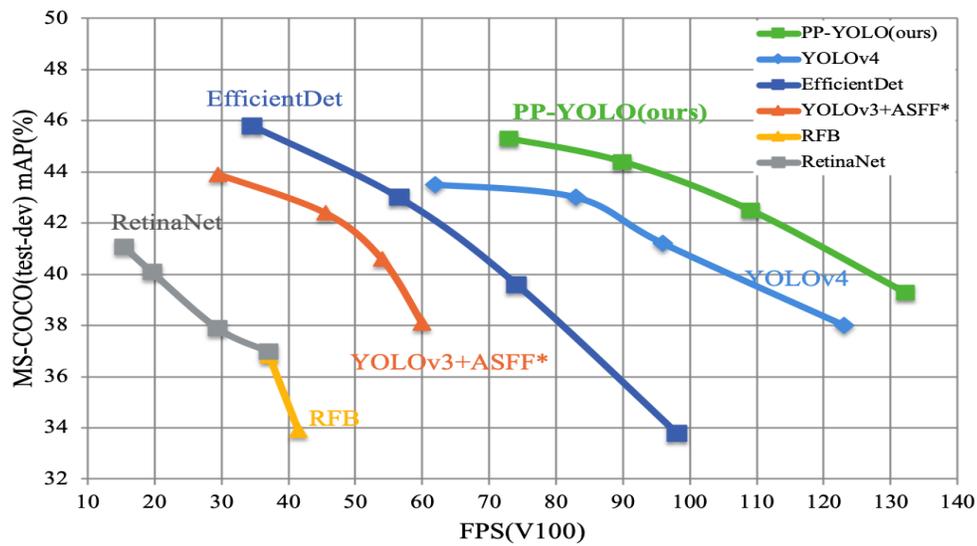

Fig 12

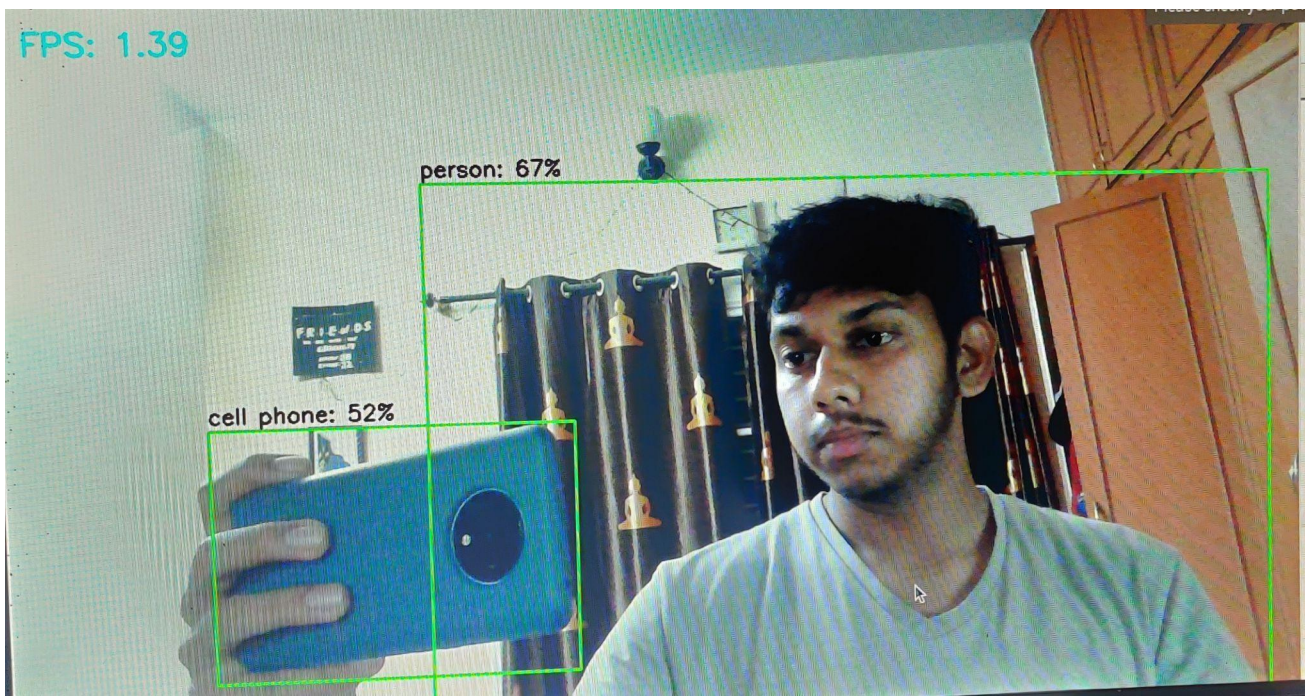

Fig 13

**OCR ENGINE COMPARISON –TESSERACT VS EASY OCR:**

1. ACCURACY: We have tested with texts and numbers of about 1000 samples. Generated random alphabets/numbers on a blank image , Tesseract and EasyOCR are used in parsing the image.

Scenario 1 : For 1000 sample images of alphabets with two words generated by random words and given input data flood experience

Scenario 2: For 1000 sample images of numbers with 5-digit and 2 decimal points numbers are used given input data 49403.65.

The detailed comparison and errors noticed:

|  | Error Rate on Numbers | Error Rate on Alphabets | Misinterpret Numbers | Misinterpret Alphabets |
|---|---|---|---|---|
| **Tesseract** | 5.50% | 0.70% | miss continuous 7<br>miss continuous 2<br>miss . | misinterpret t to r<br>miss continuous l<br>add ' in front of o<br>add . in the end<br>add , in the end |
| **EasyOCR** | 1.90% | 4.30% | miss 1 in the end<br>misinterpret . to _ | misinterpret l to i<br>misinterpret h to n<br>misinterpret f to t<br>misinterpret d to a<br>miss y in the end<br>miss v |

Fig 14

2. SPEED:for the same scenarios, below are the detailed comparisons

|  | CPU | GPU |
|---|---|---|
| **Tesseract** | 0.3 seconds/image | 0.25 seconds/image |
| **EasyOCR** | 0.82 seconds/image | 0.07 seconds/image |

Fig 15

Conclusion:
- In conclusion, tesseract does a better job in recognising alphabets and easy OCRr's for recognition of digits .
- Tesseract is more recommended on a cpu and easy ocr on a gpu as it's more quicker .

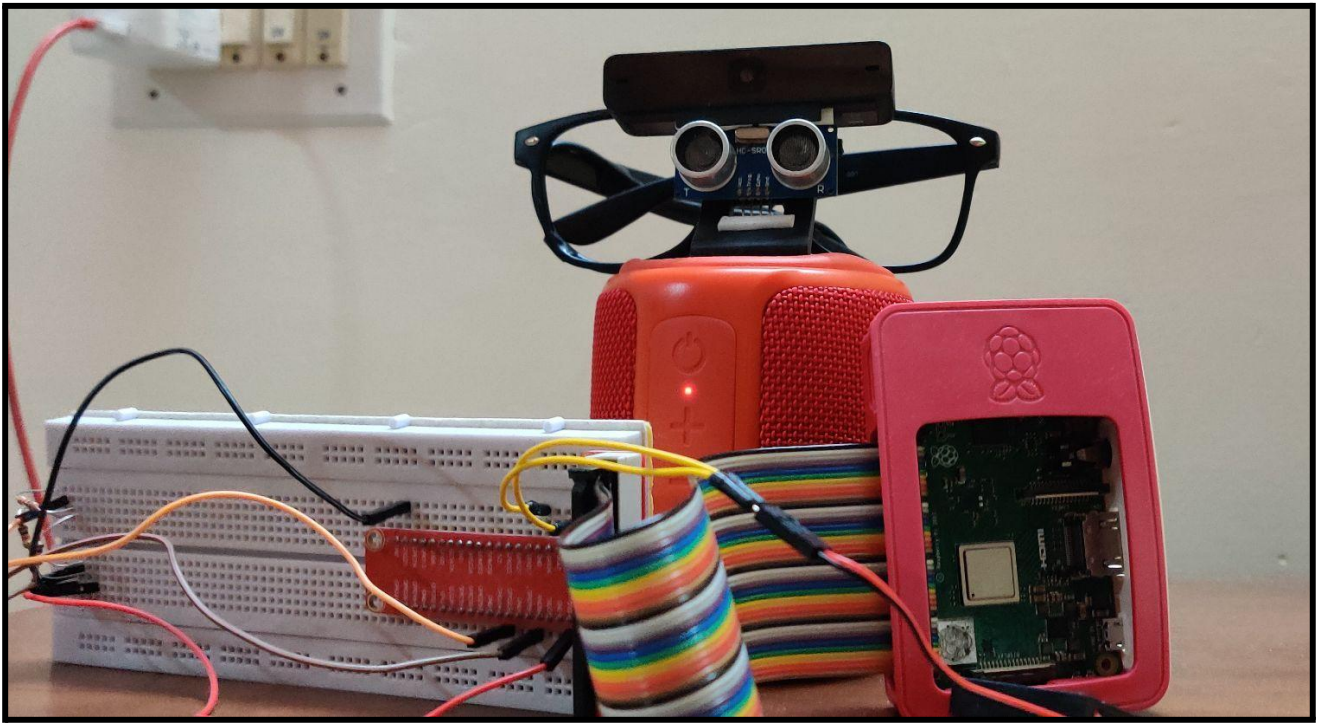

Fig 16

# Chapter 5 – Conclusion and Future work

We have put together a functional prototype of a blind stick that helps the specially disabled to travel to the places of their choice with the help of multiple services offered by our model. By incorporating models that perform proximity detection using an ultrasonic SR04 sensor, object detection using a camera and finally OCR to read the label on the recognised object. Once read, the mycroft voice assistant will read out the recognised object via a speaker which provides guidance for the user about his surroundings for that instant. Based on some preliminary tests we found the average computing time of the whole sequence of processes to be 3-5 seconds.

Here are some scopes to our work in the future :

1. Involving facial recognition technique would help the person to recognise that someone who he knows is in the surroundings.
2. When the person is stationary and has a complete description of the surroundings, we could incorporate a unique process that helps the person in doing so.
3. Emotional intelligence too is a possible scope to explore as it gives the user an idea of the person's emotion.
4. Personal real time navigation could also be a productive feature that would basically behave like google maps but with additional benefits of the pre existing features.
5. Certain VQA models could also be constructed based on the current work which inturn helps the blind to ask queries regarding the surroundings and the objects he comes across in his daily life.
6. Lie detection is another intriguing domain which the model could assimilate. This would help the user to be not exploited in his life.
7. Better processors and investments are required to be put in to accommodate all the functionalities to run hand in hand and produce the best accurate result in minimal time.
8. Various models of IOT can be implemented parallelly.
9. Embedding the GPS functionality to track the person by his loved ones.

In conclusion, our work helps blind people to get a new and wider perspective of things and their surroundings in their day to day life, making them feel more empowered and confident.